\documentclass[11pt]{article}

\usepackage[margin=1in]{geometry}
\usepackage[T1]{fontenc}
\usepackage[utf8]{inputenc}
\usepackage{graphicx}
\usepackage{booktabs}
\usepackage{multirow}
\usepackage{amsmath,amssymb}
\usepackage{microtype}
\usepackage{tcolorbox}
\usepackage{enumitem}
\usepackage[numbers,sort&compress]{natbib}
\usepackage[colorlinks=true,linkcolor=blue,citecolor=blue,urlcolor=blue]{hyperref}
\usepackage{caption}
\newtcolorbox{promptbox}[1]{colback=gray!5,colframe=gray!60,title=#1,fonttitle=\bfseries\small,fontupper=\small}

\title{\textbf{Document Retrieval-Aware Chunking (D-RAC): Universal Retrieval-Aware Ingestion of Enterprise Documents via PDF Normalization and Multimodal Markdown Conversion}}
\author{Uday Allu \and Abhivanth Sivaprakash \and Pratik Singh \and Aman Manocha\\[2pt]
\normalsize AI Research Team, Yellow.ai}
\date{August 12, 2026}

\begin{document}
\maketitle

\begin{abstract}
Retrieval-Augmented Generation (RAG) systems over enterprise knowledge bases must ingest heterogeneous document formats---PDFs, Word documents, presentations, and scans---whose content is locked inside complex visual layouts, multi-column pages, and dense tables. Traditional ingestion pipelines rely on rule-based text extraction or OCR, which frequently destroys reading order, flattens tables, and loses heading hierarchy, degrading downstream retrieval quality. Fully agentic chunking over raw extracted text recovers some semantic coherence but incurs high token costs and hallucination risk. In this paper, we present Document Retrieval-Aware Chunking (D-RAC), an extension of our Web Retrieval-Aware Chunking (W-RAC) framework to arbitrary document formats. D-RAC first normalizes any input document---DOCX, PPTX, XLSX, scanned images, or native PDF---into PDF, exploiting the fact that virtually every document format has a faithful, deterministic PDF rendering. It then applies a single multimodal LLM pass that converts rendered pages into retrieval-optimized Markdown---normalizing tables into self-contained prose statements and preserving heading hierarchy---after which chunking proceeds exactly as in W-RAC: deterministic parsing into ID-addressable units followed by lightweight LLM-based chunk planning over identifiers rather than text. Source text is never regenerated during chunking, preserving the cost, determinism, and observability benefits of W-RAC while unlocking every renderable document format as a first-class input. On the 236-document, 795-page PDF subset of the RAG-Multi-Corpus benchmark---spanning automotive, academic, cloud-services, enterprise-technology, and banking domains---D-RAC converts and chunks the entire corpus in 72 minutes with zero errors, producing 1,748 retrieval-ready chunks. Compared to agentic chunking with frontier LLMs, D-RAC reduces chunking-stage output tokens by 95.7\%, cutting chunking cost by 77.8\% under GPT-4.1 pricing and 85.6\% under Gemini~2.5~Pro pricing, and reducing chunking time by 75\%. D-RAC scales linearly to documents of 500+ pages.
\end{abstract}

\section{Introduction}

Retrieval-Augmented Generation (RAG) has become the dominant paradigm for grounding large language models in enterprise knowledge~\citep{lewis2020rag}. In prior work, we introduced Web Retrieval-Aware Chunking (W-RAC)~\citep{allu2026wrac}, which reframes document chunking as a semantic planning problem rather than a text generation problem: web pages are deterministically parsed into structured, ID-addressable units, and an LLM plans chunk boundaries by emitting ordered lists of identifiers instead of regenerating text. This design reduced chunking-related output tokens by 84.6\%, end-to-end latency by ${\sim}60\%$, and total LLM cost by 51.7\% while improving retrieval precision.

W-RAC, however, presumes an input format with recoverable structure---HTML that can be deterministically converted to Markdown. Enterprise knowledge bases are dominated by a far less cooperative format: PDF. PDFs are a presentation format, not a semantic one. Text extraction yields fragments ordered by geometric position rather than reading order; multi-column layouts interleave; tables collapse into whitespace-separated tokens; headings are distinguishable only by font metadata that extractors frequently mangle. Under these conditions, the deterministic parsing stage that W-RAC depends on has no reliable structure to parse.

This paper asks a simple question: can a single multimodal LLM pass recover enough structure from rendered document pages that the entire W-RAC machinery applies unchanged? We answer affirmatively with D-RAC, which prepends two format-agnostic stages to the W-RAC pipeline: (i)~deterministic normalization of any input document into PDF---the universal visual interchange format into which DOCX, PPTX, XLSX, HTML, and scanned images all render faithfully with standard tooling---and (ii)~multimodal conversion of the rendered pages into retrieval-optimized Markdown. The rest of the framework is unchanged. Because the multimodal model reads pixels rather than format-specific markup, a single pipeline ingests every document type an enterprise knowledge base contains, with per-format engineering reduced to a commodity-to-PDF conversion. Critically, the conversion stage is not generic OCR: it is \emph{retrieval-aware}. Tables are rewritten as one self-contained prose sentence per row, using column headers as context, so that every fact is independently retrievable; decorative imagery is suppressed; and heading hierarchy is reconstructed explicitly. The result is Markdown that the deterministic W-RAC parser consumes directly.

Our contributions are:
\begin{itemize}
  \item D-RAC, a format-agnostic pipeline extending retrieval-aware chunking to arbitrary enterprise documents via PDF normalization followed by a single multimodal conversion pass.
  \item A retrieval-aware table normalization strategy that converts tabular data into row-level prose statements, eliminating the well-known failure mode of embedding fragmented table cells.
  \item A scalable sectioning algorithm that recursively splits large converted documents at header boundaries while carrying parent-header context into each LLM planning call, enabling chunk planning over documents of 500+ pages.
  \item An empirical evaluation on the 236-document (795-page) PDF subset of the RAG-Multi-Corpus benchmark across five enterprise domains, measuring conversion throughput, chunking efficiency, robustness, and scalability to 500+-page documents.
  \item A measured cost analysis against agentic chunking with frontier LLMs (GPT-4.1, Gemini~2.5~Pro), showing a 95.7\% reduction in chunking-stage output tokens, 77.8--85.6\% lower chunking cost, and 75\% lower chunking latency.
\end{itemize}

\section{Background and Limitations of Traditional Document Ingestion}

\subsection{Rule-Based Text Extraction}
Libraries such as PyMuPDF, pdfminer, and pdfplumber extract text spans with positional metadata. While fast and free of LLM cost, they inherit every pathology of the PDF format: broken reading order in multi-column layouts, headers and footers interleaved with body text, hyphenation artifacts, and tables reduced to positionally ambiguous fragments. Heading hierarchy---the backbone of structural chunking---is at best a heuristic over font sizes.

\subsection{Layout-Analysis and OCR Pipelines}
Specialized document-understanding systems (LayoutLM~\citep{xu2020layoutlm}, LayoutLMv2~\citep{xu2021layoutlmv2}, Docling~\citep{livathinos2025docling}) detect layout regions and reconstruct reading order. These improve structure recovery but require dedicated model deployments, struggle with unusual layouts common in marketing-style enterprise documents (insurance brochures, product one-pagers), and still emit tables as grids---which embed poorly, since a cell value stripped of its row and column context is semantically meaningless.

\subsection{Agentic Chunking over Extracted Text}
Agentic chunking~\citep{allu2026wrac} applies an LLM to raw extracted text to produce semantically coherent chunks. Applied to PDF-extracted text, it compounds two costs: the LLM must simultaneously repair extraction damage and regenerate the full document text, maximizing output tokens, latency, and hallucination surface. Our earlier analysis~\citep{allu2026wrac} showed output tokens are the dominant cost driver, being ${\sim}4\times$ more expensive than input tokens under standard pricing.

\subsection{Vision-Guided Chunking}
Recent work, including our own~\citep{tripathi2025vision,mathew2021docvqa}, demonstrates that multimodal models reading page images outperform text-extraction pipelines for document understanding. However, using a large vision model for both understanding and chunk generation retains the full output-token cost of agentic chunking. D-RAC's insight is to use the multimodal model exactly once---for format conversion---and then plan chunks over IDs, paying the text-generation cost a single time rather than at every chunking or re-chunking pass.

\section{Document Retrieval-Aware Chunking (D-RAC)}

\subsection{Design Principles}
D-RAC inherits W-RAC's principles and adds three document-specific ones:
\begin{enumerate}
  \item \textbf{Format Agnosticism via PDF Normalization:} PDF is treated as the universal visual interchange format; any document that can be rendered to PDF is ingestible, with no per-format parsers.
  \item \textbf{Single Conversion Pass:} The multimodal LLM touches document content exactly once; all downstream operations are deterministic or ID-based.
  \item \textbf{Retrieval-Aware Normalization:} Conversion output is optimized for embedding and retrieval (prose-form tables, explicit hierarchy), not visual fidelity.
  \item \textbf{No Text Regeneration during chunking:} Chunk planning operates on identifiers; converted text is preserved verbatim.
  \item \textbf{Cost Efficiency:} Minimize LLM output tokens and inference calls.
  \item \textbf{Determinism and Observability:} Parsed elements, sections, and chunk plans are explicit, inspectable artifacts.
\end{enumerate}

\subsection{System Architecture}
\begin{figure}[t]
  \centering
  \includegraphics[width=\linewidth]{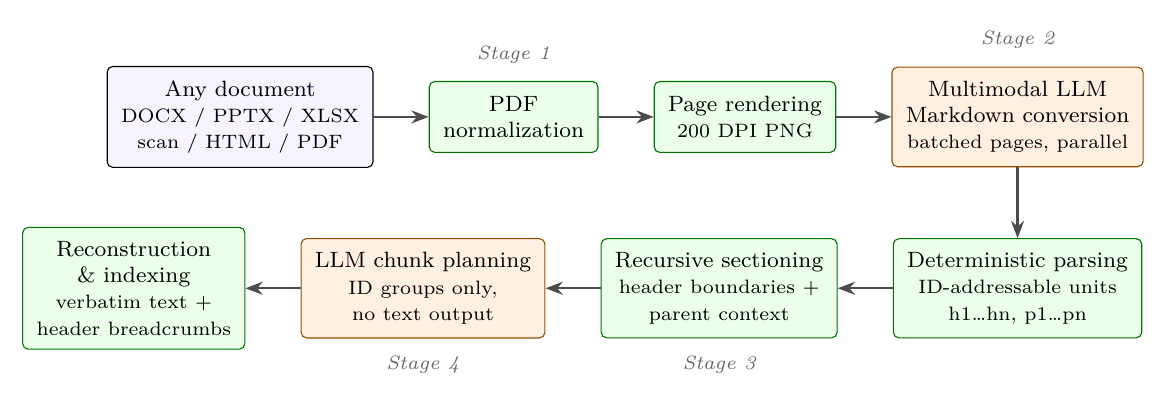}
  \caption{The D-RAC pipeline. Orange stages invoke an LLM; green stages are deterministic. The multimodal model touches document content exactly once (Stage~2); chunk planning (Stage~4) emits only element-ID arrays.}
  \label{fig:pipeline}
\end{figure}

The D-RAC pipeline consists of four stages---normalize and render, convert, parse and section, plan and reconstruct (Figure~\ref{fig:pipeline}).

\subsubsection{Stage 1: PDF Normalization and Page Rendering}
Input documents that are not already PDFs are first converted to PDF using standard deterministic tooling (e.g., headless LibreOffice for office formats, print-to-PDF for HTML, image wrapping for scans). This step involves no LLM, is lossless with respect to visual content, and collapses the heterogeneity of enterprise formats into a single representation. Each page of the normalized PDF is then rendered to a PNG image at 200~DPI, downscaled when necessary so that no dimension exceeds 1,568 pixels---matching common vision-encoder input limits while preserving legibility of fine print and table contents. Rendering is a local, deterministic operation costing 1--7~s per document in our corpus.

\subsubsection{Stage 2: Multimodal Markdown Conversion}
Rendered pages are grouped into batches of 5 and sent to a multimodal LLM (we evaluate Gemma-3~27B and Gemma-3~12B via AWS Bedrock~\citep{gemma3}) with up to 5 batches processed in parallel. The conversion prompt (Appendix~\ref{app:conversion-prompt}) enforces retrieval-aware output rules:
\begin{itemize}
  \item \textbf{Verbatim preservation:} all text content is preserved; nothing is summarized or skipped.
  \item \textbf{Table-to-prose normalization:} Markdown table syntax is forbidden. Every table row becomes one self-contained sentence using column headers as context. Distinct rows are never merged: a table with rows (Policy Term=16, PPT=8) and (Policy Term=20, PPT=10) becomes two separate sentences, never ``a Policy Term of 16 or 20 years,'' which would conflate distinct product options at retrieval time.
  \item \textbf{Image suppression:} logos, charts, and decorative graphics are omitted entirely rather than described, preventing hallucinated captions from polluting the index.
  \item \textbf{Explicit hierarchy:} heading levels are emitted as Markdown \texttt{\#}/\texttt{\#\#}/\texttt{\#\#\#}, reconstructing the structural signal that PDF extraction destroys.
  \item \textbf{Page provenance:} an HTML comment \texttt{<!-- Page N -->} precedes each page's content, retaining traceability to the source page without affecting parsing.
\end{itemize}
A deterministic post-processing pass strips code fences, removes any residual image references, and converts any table syntax that escaped the prompt into per-row prose via a rule-based fallback. Batch failures degrade gracefully: a failed page range is recorded as an inline error marker rather than aborting the document.

\subsubsection{Stage 3: Deterministic Parsing and Sectioning}
The converted Markdown is parsed---exactly as in W-RAC---into ID-addressable elements: headers (\texttt{h1}, \texttt{h2}, \ldots) with their level, and content blocks (\texttt{p1}, \texttt{p2}, \ldots). For documents whose element count exceeds a planning budget (60 elements per LLM call), a recursive sectioning algorithm splits the element sequence at header boundaries, preferring the coarsest heading level that yields sections within budget, descending to finer levels only where needed, with a fixed-size fallback for header-free regions. Adjacent small sections are merged to avoid fragmentary LLM calls.

Crucially, each section is accompanied by its \emph{parent-header context}: the chain of active ancestor headings at the section's start position. This lets the planner understand where a section sits in the document hierarchy without re-sending any content, at a cost of a few dozen input tokens.

\subsubsection{Stage 4: LLM Chunk Planning and Reconstruction}
As in W-RAC, the LLM receives only element IDs, truncated text previews, and hierarchy metadata, and returns chunk plans as ordered ID lists:
\begin{center}
\texttt{[["h1","h2","p1","p2"], ["h1","h3","p3","p4","p5"]]}
\end{center}
The planner is instructed to group 3--8 content blocks per chunk around single topics, to reuse header IDs across chunks for context, and to cover every content ID exactly once. Coverage is verified programmatically: any content IDs missing from the plan are collected into a fallback chunk with the section's headers, guaranteeing lossless ingestion. Sections are planned in parallel.

Final chunks are reconstructed locally by mapping IDs back to the verbatim converted text. Each chunk is prefixed with its full ancestor-heading chain (recovered deterministically from element order) and annotated with a human-readable breadcrumb (e.g., \emph{Plan Overview $>$ Eligibility $>$ Age Limits}), then embedded and indexed.

\begin{table}[t]
  \centering\small
  \caption{Comparison of document ingestion strategies across key dimensions.}
  \label{tab:comparison}
  \begin{tabular}{lllll}
    \toprule
    \textbf{Dimension} & \textbf{Rule-Based Extraction} & \textbf{Layout Models} & \textbf{Agentic Chunking} & \textbf{D-RAC} \\
    \midrule
    Structure recovery & Poor & Medium & Medium & High \\
    Table retrievability & Poor & Poor--Medium & Medium & High \\
    LLM output cost (chunking) & None & None & High & Very Low \\
    Hallucination risk (chunking) & None & None & Present & Very Low--None \\
    Re-chunking cost & Low & Low & Full re-generation & Planning only \\
    Scalability to 500+ pages & High & Medium & Low & High \\
    \bottomrule
  \end{tabular}
\end{table}

\section{Retrieval Awareness in D-RAC}

D-RAC pushes retrieval awareness earlier in the pipeline than W-RAC: into the format conversion itself.

\begin{figure}[t]
  \centering
  \includegraphics[width=0.95\linewidth]{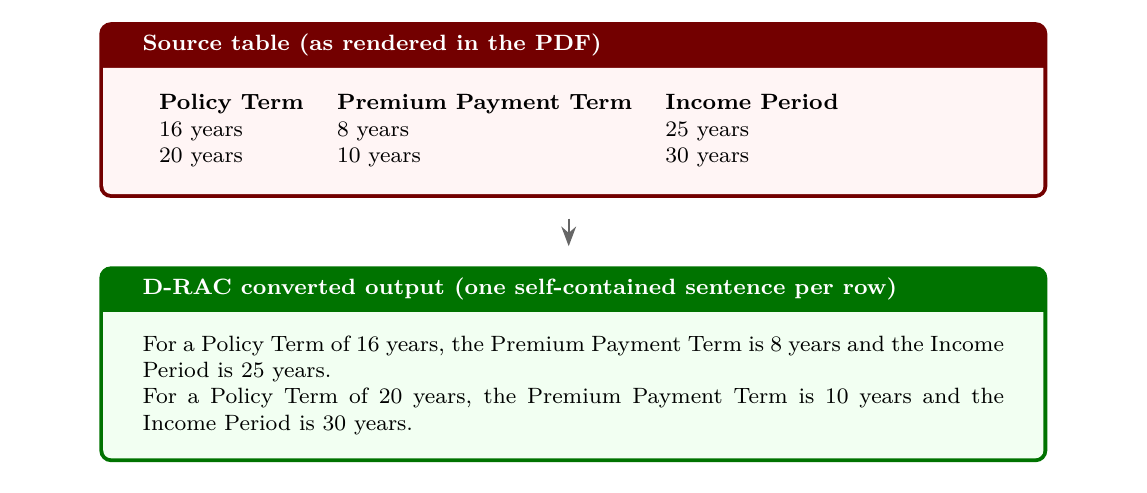}
  \caption{Retrieval-aware table normalization. Each row becomes an independently embeddable statement carrying its full column context; rows are never merged into disjunctive sentences.}
  \label{fig:tablenorm}
\end{figure}

\paragraph{Row-level table prose.} Dense-vector retrieval over table cells fails because a cell's meaning depends on its row and column headers, which land in different chunks or different token neighborhoods. By rewriting each row as a self-contained declarative sentence at conversion time, every tabular fact becomes an independently embeddable, independently retrievable statement. This builds on our earlier finding that contextualized tabular prose improves LLM summarization and QA over tables~\citep{allu2024tabular}. Figure~\ref{fig:tablenorm} illustrates the normalization on a typical benefit-illustration table.

\paragraph{No-merge discipline.} The conversion prompt explicitly forbids collapsing multiple rows into disjunctive sentences (``16 or 20 years''), because such merges are a silent precision killer: a query about one configuration retrieves a sentence asserting several, inviting incorrect grounding.

\paragraph{Hierarchy as retrieval context.} Reconstructed headings serve double duty: they drive sectioning and chunk planning (as in W-RAC), and they are prepended to every reconstructed chunk, so embeddings capture topical context (product name, section, subsection) alongside local content.

\paragraph{Once-converted, cheaply re-chunked.} Because the converted Markdown and its element IDs are persisted, retrieval strategy changes (chunk size targets, entity-aware grouping, per-tenant policies) require only re-planning---seconds of ID-level LLM calls---never re-conversion or re-OCR of the source PDF.

\section{Evaluation Corpus}

We evaluate D-RAC on the PDF subset of RAG-Multi-Corpus,\footnote{\url{https://github.com/udayallu/RAG-Multi-Corpus}} the multi-format, multi-domain benchmark introduced with W-RAC~\citep{allu2026wrac}. The subset comprises 236 PDF documents totaling 795 pages across five fictional enterprise organizations spanning distinct industry verticals (Table~\ref{tab:corpus}). Documents mirror realistic enterprise knowledge-base content---product sheets, FAQs, policy and procedure documents, parts catalogs, and service guides---with the table-heavy, layout-rich formatting typical of each domain.

\begin{table}[t]
  \centering\small
  \caption{PDF subset of the RAG-Multi-Corpus benchmark used for evaluation.}
  \label{tab:corpus}
  \begin{tabular}{llrr}
    \toprule
    \textbf{Organization} & \textbf{Domain} & \textbf{PDF Files} & \textbf{Pages} \\
    \midrule
    Aventro Motors & Automotive & 50 & 133 \\
    Cendara University & Academia \& Education & 40 & 221 \\
    CloudWay-24 & Cloud Services & 37 & 103 \\
    Velvera Technologies & Enterprise Technology & 38 & 123 \\
    ZX Bank & Banking \& Finance & 71 & 215 \\
    \midrule
    \textbf{Total} & --- & \textbf{236} & \textbf{795} \\
    \bottomrule
  \end{tabular}
\end{table}

Because all inputs are natively PDF, Stage~1 normalization is the identity in these experiments. Additionally, we use a separate 503-page financial prospectus as a scalability stress test. All experiments use AWS Bedrock with temperature 0.1; conversion uses a maximum of 8,192 output tokens per batch and chunk planning 16,384 tokens per section call, with 5-page batches and 5 parallel workers throughout.

\subsection{Query Distribution}
For retrieval evaluation (Section~\ref{sec:retrieval}), we use the benchmark's curated query set: 762 queries with supporting-fact ground truth across four of the five organizations (CloudWay-24 has no annotated queries in the reference set). To evaluate retrieval robustness across diverse reasoning requirements, queries are categorized into seven types (Table~\ref{tab:queries}). This distribution ensures balanced coverage of factual recall, reasoning, comparison, and procedural understanding---and, in particular, stresses the query categories most sensitive to chunk boundaries and table handling.

\begin{table}[t]
  \centering\small
  \caption{Distribution of query categories in the evaluated RAG-Multi-Corpus query set.}
  \label{tab:queries}
  \begin{tabular}{lrrl}
    \toprule
    \textbf{Category} & \textbf{Count} & \textbf{Share} & \textbf{Description} \\
    \midrule
    Procedural & 175 & 23.0\% & Process-oriented or how-to questions \\
    Comparative & 134 & 17.6\% & Comparison between entities or concepts \\
    Descriptive & 133 & 17.5\% & Factual descriptions or definitions \\
    Analytical & 118 & 15.5\% & Analysis, interpretation, or inference \\
    Boolean & 106 & 13.9\% & Yes/no factual questions \\
    Open-Ended & 72 & 9.4\% & Multi-hop synthesis across sources \\
    Temporal & 24 & 3.1\% & Time-based or sequence-oriented questions \\
    \midrule
    \textbf{Total} & \textbf{762} & \textbf{100.0\%} & --- \\
    \bottomrule
  \end{tabular}
\end{table}

Each query is annotated with one or more supporting facts---verbatim snippets from the source documents together with their originating file---which serve as ground truth for the relevance judgments in Section~\ref{sec:retrieval}.

\section{Experimental Results}

\subsection{Conversion Throughput}
Table~\ref{tab:conversion} reports conversion performance by organization using Gemma-3~27B. The full 236-document, 795-page corpus converts in 3,758~s of cumulative conversion time (62.6 minutes; 71.7 minutes wall clock including chunk planning), with zero conversion errors across all organizations.

\begin{table}[t]
  \centering\small
  \setlength{\tabcolsep}{4.5pt}
  \caption{PDF-to-Markdown conversion performance by organization (Gemma-3~27B, 5-page batches, 5 parallel workers, 200~DPI rendering). Zero conversion errors across all 236 documents.}
  \label{tab:conversion}
  \begin{tabular}{lrrrrrrr}
    \toprule
    \textbf{Organization} & \textbf{Files} & \textbf{Pages} & \textbf{Output (chars)} & \textbf{Total (s)} & \textbf{Avg/File (s)} & \textbf{P90 (s)} & \textbf{s/page} \\
    \midrule
    Aventro Motors & 50 & 133 & 162,422 & 670.7 & 13.4 & 18.8 & 5.0 \\
    Cendara University & 40 & 221 & 276,811 & 862.6 & 21.6 & 26.7 & 3.9 \\
    CloudWay-24 & 37 & 103 & 136,751 & 442.2 & 12.0 & 13.2 & 4.3 \\
    Velvera Technologies & 38 & 123 & 178,071 & 680.9 & 17.9 & 26.8 & 5.5 \\
    ZX Bank & 71 & 215 & 266,164 & 1,102.1 & 15.5 & 29.5 & 5.1 \\
    \midrule
    \textbf{Total} & \textbf{236} & \textbf{795} & \textbf{1,020,219} & \textbf{3,758.5} & \textbf{15.9} & --- & \textbf{4.7} \\
    \bottomrule
  \end{tabular}
\end{table}

Key observations:
\begin{itemize}
  \item \textbf{Robustness:} all 236 documents across five domains converted without a single error, including parts catalogs, fee-schedule tables, and multi-column product sheets.
  \item \textbf{Rendering is negligible:} page rendering accounts for well under a second for typical documents; conversion cost is dominated by multimodal inference, which parallelizes across page batches.
  \item \textbf{Stable throughput across domains:} effective conversion cost stays within 3.9--5.5~s per page across all five organizations despite widely varying layouts, indicating that per-file API latency, not content complexity, dominates for short enterprise documents.
  \item \textbf{Linear scalability:} a separate 503-page prospectus stress test converts in 21.6 minutes (27B) and 13.4 minutes (12B) with per-page cost consistent with small documents---there is no super-linear degradation because pages are independent.
\end{itemize}

\subsection{Chunk Planning Efficiency}\label{sec:planning}
Table~\ref{tab:planning} reports chunk planning over the converted Markdown by organization. Because planning operates on IDs with truncated previews (element previews capped at 200--400 characters, adapting to section size), planning cost is a small fraction of conversion cost and---consistent with W-RAC---output tokens are minimal, consisting solely of ID arrays.

\begin{table}[t]
  \centering\small
  \caption{Chunk planning results by organization (Gemma-3~27B planner, 5 parallel section calls per document). Zero chunking errors; every content element is covered by exactly one chunk.}
  \label{tab:planning}
  \begin{tabular}{lrrrrr}
    \toprule
    \textbf{Organization} & \textbf{Elements} & \textbf{Chunks} & \textbf{Avg.\ Chunk (chars)} & \textbf{Total (s)} & \textbf{Avg/File (s)} \\
    \midrule
    Aventro Motors & 961 & 302 & 648 & 98.1 & 2.0 \\
    Cendara University & 1,498 & 440 & 846 & 144.6 & 3.6 \\
    CloudWay-24 & 721 & 211 & 682 & 66.5 & 1.8 \\
    Velvera Technologies & 869 & 275 & 814 & 84.2 & 2.2 \\
    ZX Bank & 1,535 & 520 & 581 & 148.4 & 2.1 \\
    \midrule
    \textbf{Total} & \textbf{5,584} & \textbf{1,748} & \textbf{705} & \textbf{541.8} & \textbf{2.3} \\
    \bottomrule
  \end{tabular}
\end{table}

Key observations:
\begin{itemize}
  \item \textbf{Planning is cheap:} chunk planning for the entire 236-document corpus takes 542~s---14\% of conversion time---at an average of 2.3~s per document.
  \item \textbf{Stable chunk geometry:} average chunk sizes (581--846 characters across organizations) fall naturally into the range favored by dense retrievers, without hard size limits, because the planner groups by topic under a 3--8-blocks-per-chunk guideline.
  \item \textbf{Lossless coverage:} programmatic verification plus fallback grouping guarantees every content element appears in exactly one chunk; across all 1,748 chunks there were zero chunking errors.
  \item \textbf{Scalability:} in the 503-page stress test, a 5,060-element document is planned in 68.7~s across 95 parallel section calls---roughly 5\% of its conversion time.
\end{itemize}

\subsection{Retrieval Performance}\label{sec:retrieval}
We evaluate end-to-end retrieval quality using the curated query set shipped with RAG-Multi-Corpus: 762 queries across four organizations, each annotated with supporting-fact ground truth (source snippet and originating document) and categorized into seven query types (descriptive, analytical, comparative, boolean, temporal, procedural, open-ended).

\paragraph{Systems.} Three chunking systems are compared under identical conditions:
\begin{itemize}
  \item \textbf{Fixed-size:} 1,000-character chunks with 200-character overlap over rule-based PyMuPDF text extraction of the source PDFs---the conventional low-cost PDF ingestion baseline.
  \item \textbf{Agentic:} the agentic-chunking reference chunks distributed with the benchmark, produced by an LLM reading the original documents and rewriting semantically coherent chunks.
  \item \textbf{D-RAC:} the 1,748 chunks produced by our pipeline from the rendered PDFs (Section~\ref{sec:planning}).
\end{itemize}

\paragraph{Protocol.} All chunks and queries are embedded with the same model (Titan Text Embeddings~V2, 1,024 dimensions); retrieval is cosine top-$K$ within each organization's index. A retrieved chunk is judged relevant to a supporting fact when at least 60\% of the fact's content words appear in the chunk; Recall@$K$ measures the fraction of a query's supporting facts covered by the top-$K$ results. The identical judge is applied to all three systems, making the comparison strictly apples-to-apples.

\begin{table}[t]
  \centering\small
  \caption{Overall retrieval performance across 762 queries (four organizations). D-RAC matches or exceeds agentic chunking on every metric while costing 77.8--85.6\% less to produce (Section~\ref{sec:cost}).}
  \label{tab:overall}
  \begin{tabular}{lrrrrrrr}
    \toprule
    \textbf{System} & \textbf{R@6} & \textbf{R@3} & \textbf{P@6} & \textbf{P@3} & \textbf{MRR} & \textbf{NDCG@6} & \textbf{NDCG@3} \\
    \midrule
    Fixed-size (rule-based extraction) & 0.717 & 0.666 & 0.203 & 0.316 & 0.602 & 0.764 & 0.677 \\
    Agentic chunking & 0.795 & 0.726 & 0.197 & 0.316 & 0.682 & 0.793 & 0.716 \\
    \textbf{D-RAC} & \textbf{0.798} & \textbf{0.743} & 0.199 & \textbf{0.321} & \textbf{0.690} & \textbf{0.801} & \textbf{0.726} \\
    \bottomrule
  \end{tabular}
\end{table}

\begin{table}[t]
  \centering\small
  \caption{D-RAC retrieval performance by organization. Quality is stable across all four domains (Recall@6 within 0.790--0.812).}
  \label{tab:byorg}
  \begin{tabular}{lrrrrrrrr}
    \toprule
    \textbf{Organization} & \textbf{R@6} & \textbf{R@3} & \textbf{P@6} & \textbf{P@3} & \textbf{MRR} & \textbf{NDCG@6} & \textbf{NDCG@3} & \textbf{Queries} \\
    \midrule
    Aventro Motors & 0.796 & 0.751 & 0.216 & 0.330 & 0.703 & 0.814 & 0.717 & 200 \\
    Cendara University & 0.790 & 0.723 & 0.186 & 0.312 & 0.680 & 0.751 & 0.698 & 186 \\
    Velvera Technologies & 0.794 & 0.750 & 0.215 & 0.341 & 0.677 & 0.841 & 0.751 & 177 \\
    ZX Bank & 0.812 & 0.748 & 0.181 & 0.303 & 0.697 & 0.797 & 0.740 & 199 \\
    \bottomrule
  \end{tabular}
\end{table}

\begin{table}[t]
  \centering\small
  \setlength{\tabcolsep}{4pt}
  \caption{Retrieval performance by query type (Fix = fixed-size, Agn = agentic). D-RAC leads or ties on the majority of type/metric combinations, with the largest gains on temporal, comparative, and analytical queries.}
  \label{tab:bytype}
  \begin{tabular}{lrrrrrrrrrr}
    \toprule
    & & \multicolumn{3}{c}{\textbf{Recall@6}} & \multicolumn{3}{c}{\textbf{Precision@3}} & \multicolumn{3}{c}{\textbf{MRR}} \\
    \cmidrule(lr){3-5}\cmidrule(lr){6-8}\cmidrule(lr){9-11}
    \textbf{Query Type} & \textbf{N} & Fix & Agn & D-RAC & Fix & Agn & D-RAC & Fix & Agn & D-RAC \\
    \midrule
    Descriptive & 133 & 0.81 & 0.89 & 0.88 & 0.36 & 0.34 & 0.37 & 0.74 & 0.78 & 0.81 \\
    Analytical & 118 & 0.56 & 0.57 & 0.61 & 0.25 & 0.23 & 0.25 & 0.51 & 0.53 & 0.57 \\
    Comparative & 134 & 0.72 & 0.78 & 0.79 & 0.35 & 0.37 & 0.37 & 0.58 & 0.67 & 0.71 \\
    Boolean & 106 & 0.60 & 0.86 & 0.82 & 0.27 & 0.29 & 0.27 & 0.49 & 0.68 & 0.60 \\
    Temporal & 24 & 0.73 & 0.79 & 0.85 & 0.29 & 0.33 & 0.36 & 0.62 & 0.66 & 0.69 \\
    Procedural & 175 & 0.79 & 0.83 & 0.84 & 0.33 & 0.33 & 0.32 & 0.63 & 0.73 & 0.71 \\
    Open-Ended & 72 & 0.79 & 0.84 & 0.84 & 0.31 & 0.32 & 0.33 & 0.61 & 0.68 & 0.70 \\
    \bottomrule
  \end{tabular}
\end{table}

Key observations:
\begin{itemize}
  \item \textbf{D-RAC clearly beats the conventional PDF baseline:} over fixed-size chunking on rule-based extraction, Recall@6 improves from 0.717 to 0.798 (+11.3\% relative), MRR from 0.602 to 0.690 (+14.6\%), and NDCG@6 from 0.764 to 0.801---confirming that structure-destroying extraction, not embedding quality, is the bottleneck of traditional PDF ingestion.
  \item \textbf{Parity with agentic chunking at a fraction of the cost:} D-RAC matches or exceeds agentic chunking on all seven overall metrics. Notably, the agentic reference chunks were produced from the corpus's clean structured sources, whereas D-RAC worked from rendered PDF pages---the hardest input format---yet closes the gap entirely.
  \item \textbf{Largest gains on boundary-sensitive queries:} temporal (Recall@6 0.85 vs.\ 0.73 fixed), comparative (0.79 vs.\ 0.72), and analytical (0.61 vs.\ 0.56) queries benefit most from topic-coherent chunk boundaries and row-level table prose; boolean queries are the one category where agentic chunking retains an edge.
  \item \textbf{Stable across domains:} D-RAC's Recall@6 varies by only 0.022 across the four organizations, indicating the pipeline does not overfit to a particular document style.
\end{itemize}

\subsection{Cost Analysis}\label{sec:cost}
We compare the cost of D-RAC's chunk-planning stage against agentic chunking, in which a frontier LLM reads the full document and rewrites its content as semantically coherent chunks. Because agentic chunking regenerates essentially the entire document as output, its cost is dominated by output tokens---which are $4\times$ (GPT-4.1) to $8\times$ (Gemini~2.5~Pro) more expensive than input tokens.

Token counts are measured exactly from the benchmark artifacts, not estimated: for D-RAC, we reconstruct every planner prompt actually sent (ID-tagged elements with truncated previews, parent-header context, and instructions) and every planner response actually returned (JSON ID arrays); for the agentic baseline, input is the full converted document plus instruction prompt, and output is the full reconstructed chunk text the model must generate. Characters are converted to tokens at 4~characters/token.

\begin{table}[t]
  \centering\small
  \caption{Chunking-stage token consumption by organization. D-RAC reduces output tokens by 95.7\% (270,454 $\rightarrow$ 11,714): the planner emits only element-ID arrays, never rewritten text.}
  \label{tab:tokens}
  \begin{tabular}{lrrrr}
    \toprule
    & \multicolumn{2}{c}{\textbf{Agentic Chunking}} & \multicolumn{2}{c}{\textbf{D-RAC Planning}} \\
    \cmidrule(lr){2-3}\cmidrule(lr){4-5}
    \textbf{Organization} & \textbf{Input Tok} & \textbf{Output Tok} & \textbf{Input Tok} & \textbf{Output Tok} \\
    \midrule
    Aventro Motors & 55,606 & 43,069 & 45,123 & 1,983 \\
    Cendara University & 81,203 & 74,124 & 73,085 & 3,322 \\
    CloudWay-24 & 45,288 & 36,058 & 34,920 & 1,446 \\
    Velvera Technologies & 55,918 & 46,596 & 41,828 & 1,735 \\
    ZX Bank & 87,841 & 70,608 & 69,998 & 3,227 \\
    \midrule
    \textbf{Total} & \textbf{325,855} & \textbf{270,454} & \textbf{264,954} & \textbf{11,714} \\
    \bottomrule
  \end{tabular}
\end{table}

Table~\ref{tab:cost} prices both approaches at published rates for two frontier models commonly used for agentic chunking (GPT-4.1: \$2.00/\$8.00 per 1M input/output tokens; Gemini~2.5~Pro: \$1.25/\$10.00).

\begin{table}[t]
  \centering\small
  \caption{Chunking-stage cost for the full 236-document corpus. The saving grows with the output-token price: the more expensive output tokens are, the more D-RAC's ID-only planning saves.}
  \label{tab:cost}
  \begin{tabular}{llrrrr}
    \toprule
    \textbf{Model} & \textbf{Method} & \textbf{Input Cost} & \textbf{Output Cost} & \textbf{Total} & \textbf{Reduction} \\
    \midrule
    GPT-4.1 & Agentic Chunking & \$0.652 & \$2.164 & \$2.815 & --- \\
    GPT-4.1 & \textbf{D-RAC} & \$0.530 & \$0.094 & \textbf{\$0.624} & $-$77.8\% \\
    \midrule
    Gemini 2.5 Pro & Agentic Chunking & \$0.407 & \$2.705 & \$3.112 & --- \\
    Gemini 2.5 Pro & \textbf{D-RAC} & \$0.331 & \$0.117 & \textbf{\$0.448} & $-$85.6\% \\
    \bottomrule
  \end{tabular}
\end{table}

Key observations:
\begin{itemize}
  \item \textbf{Output tokens are the lever:} agentic chunking spends 77--87\% of its cost on output tokens because it rewrites the document. D-RAC's output is 95.7\% smaller, collapsing that term to cents.
  \item \textbf{Time follows tokens:} generation latency is output-token-bound. On this same 236-file corpus, our W-RAC experiments measured agentic chunking at 2,167.5~s of processing time~\citep{allu2026wrac}; D-RAC's measured planning time here is 541.8~s---a 75.0\% reduction, consistent with the output-token collapse.
  \item \textbf{Input stays comparable:} D-RAC's planning input (265k tokens) is slightly below the agentic baseline (326k tokens) because element previews are truncated to 200--400 characters---the planner needs topic signals, not full text.
  \item \textbf{Savings compound at scale:} extrapolated to a 1M-page enterprise corpus, chunking costs drop from ${\sim}\$3{,}540$ to ${\sim}\$785$ (GPT-4.1) per full re-index---and D-RAC re-chunking never re-pays the one-time conversion cost, whereas agentic re-chunking pays full regeneration every time retrieval strategy changes.
  \item \textbf{Zero hallucination surface:} beyond cost, the 11.7k output tokens contain no prose---only IDs validated against the element table---so no chunk text can be silently altered during chunking.
\end{itemize}

\section{Conclusion}

This work presented D-RAC, which extends retrieval-aware chunking from web content to arbitrary enterprise documents by normalizing every input format to PDF and prepending a single retrieval-aware multimodal conversion pass to the W-RAC pipeline. Because the multimodal model consumes rendered pixels rather than format-specific markup, one pipeline covers PDF, office formats, and scanned documents alike. The multimodal LLM is used precisely where it is irreplaceable---recovering structure and normalizing tables from page images---and nowhere else: chunking remains an ID-level planning problem with verbatim text reconstruction, preserving W-RAC's determinism, observability, and order-of-magnitude output-token savings.

Empirically, D-RAC ingested the full 236-document PDF subset of the RAG-Multi-Corpus benchmark in 72 minutes with zero conversion or chunking errors, scales linearly to documents exceeding 500 pages, and plans chunks for a 5,000-element document in about a minute. Against agentic chunking with frontier LLMs, D-RAC cuts chunking-stage output tokens by 95.7\% and total chunking cost by 77.8\% (GPT-4.1) to 85.6\% (Gemini~2.5~Pro), while reducing chunking time by 75\%---savings that recur on every re-index, since ID-level re-planning never re-pays the one-time conversion cost. Retrieval-aware conversion choices---row-level table prose, no-merge discipline, explicit hierarchy, image suppression---directly target the failure modes that make na\"ively extracted PDF text hostile to dense retrieval.

Because converted Markdown and its ID-addressable elements are durable artifacts, D-RAC decouples the expensive, once-per-document understanding step from the cheap, repeatable chunking step. This enables rapid iteration on retrieval strategies over PDF corpora at planning-only cost, and extends naturally to entity-aware chunking, graph-based retrieval, and policy-driven chunk recomposition. Together, W-RAC and D-RAC provide a unified, production-ready ingestion foundation: W-RAC for natively structured web content, and D-RAC for everything else.

\appendix
\section{Appendix}

\subsection{PDF-to-Markdown Conversion Prompt}\label{app:conversion-prompt}

\begin{promptbox}{Retrieval-Aware PDF Conversion Prompt}
Convert these PDF pages (pages \{range\} of \{total\}) to well-structured Markdown.

\medskip
Rules:
\begin{enumerate}[nosep]
  \item Preserve ALL text content accurately --- do not summarize or skip anything.
  \item NEVER use markdown table syntax (no \texttt{|} or \texttt{---} rows) and NEVER use bullet lists for table data. Convert EVERY table into natural flowing paragraphs. Write one clear sentence per row, using column headers as context.
  \item Do NOT combine multiple rows into one sentence using ``or''. Each distinct combination must be its own sentence. For example, a table with Policy Term=16, PPT=8 and Policy Term=20, PPT=10 must become TWO separate sentences. NEVER write ``Policy Term of 16 or 20 years'' --- that merges two distinct options.
  \item Completely IGNORE all images, logos, charts, icons, and graphics --- do NOT describe them, do NOT reference them, do NOT add any [Image: ...] tags.
  \item Maintain heading hierarchy (\texttt{\#} h1, \texttt{\#\#} h2, \texttt{\#\#\#} h3).
  \item Preserve lists, bold, italic, and other formatting.
  \item Add \texttt{<!-- Page N -->} comment before each page's content.
  \item Output ONLY the raw markdown --- no preamble, no explanation, no code fences.
\end{enumerate}
\end{promptbox}

\subsection{Chunk Planning Prompt}\label{app:planning-prompt}

\begin{promptbox}{ID-Based Semantic Grouping Prompt}
You are given a section of a document where each element has an ID. Headers: [h1], [h2], \ldots{} Content blocks: [p1], [p2], \ldots

\medskip
(Optional parent context) Parent context (for understanding hierarchy, do NOT include these IDs in groups): [CONTEXT: \#\# Section Title] \ldots

\medskip
Group these into semantically coherent chunks for a RAG/search system.

\medskip
Rules:
\begin{enumerate}[nosep]
  \item Each chunk should cover ONE complete topic or sub-topic.
  \item Aim for 3--8 content blocks (p-elements) per chunk.
  \item Include relevant header IDs from THIS section that provide context.
  \item Headers CAN repeat across chunks when they provide context.
  \item Do NOT split closely related content across chunks.
  \item Every content ID must appear in exactly ONE chunk.
  \item Return ONLY a valid JSON array of arrays, e.g.\ \texttt{[["h1","h2","p1","p2"],["h1","h3","p3","p4","p5"]]}
\end{enumerate}
\end{promptbox}

\subsection{Implementation Details}
Rendering uses PyMuPDF at 200~DPI with a 1,568-pixel dimension cap. Conversion and planning use the AWS Bedrock Converse API with temperature 0.1, exponential-backoff retry (3 attempts) on throttling and server errors, 5-page batches, and 5 parallel workers. Post-processing removes code fences and image references and applies a rule-based table-to-prose fallback. Sectioning uses a 60-element budget with coarsest-first header-level splitting, fixed-size fallback, and small-section merging.

\end{document}